%% file: collas2023_conference.tex
\documentclass{article} 
\usepackage{collas2023_conference,times}
\usepackage{easyReview}

\input{math_commands.tex}

\usepackage{hyperref}
\hypersetup{
    colorlinks=true,
    linkcolor=red,
    filecolor=magenta,
    urlcolor=blue,
    citecolor=purple,
    pdfpagemode=FullScreen,
    }

\usepackage{xcolor}
\usepackage{paralist}
\usepackage{caption}
\usepackage{subcaption}
\usepackage{amsfonts}
\usepackage{amsmath}
\usepackage{upgreek}
\usepackage{bm}

\newcommand\Ex{\mathbb{E}}
\newcommand\sparho{\Check{\rho}}

\newcommand\sparhov{\bm{\Check{\uprho}}}
\newcommand\muv{\bm{\upmu}}
\newcommand\piv{\bm{\uppi}}
\newcommand\Qv{\mathbf{Q}}
\newcommand\onev{\mathbf{1}}
\newcommand\cv{\mathbf{c}}

\title{Value-aware Importance Weighting for Off-policy Reinforcement Learning}

\author{Kristopher De Asis\\
Department of Computing Science\\
University of Alberta\\
Edmonton, Canada\\
\texttt{kldeasis@ualberta.ca} \And
Eric Graves\\
Department of Computing Science\\
University of Alberta\\
Edmonton, Canada\\
\texttt{graves@ualberta.ca} \And
Richard S. Sutton\\
Department of Computing Science\\
University of Alberta\\
Edmonton, Canada\\
\texttt{rsutton@ualberta.ca}
}

\collasfinalcopy 

\begin{document}

\maketitle

\begin{abstract}
Importance sampling is a central idea underlying off-policy prediction in reinforcement learning. It provides a strategy for re-weighting samples from a distribution to obtain unbiased estimates under another distribution. However, importance sampling weights tend to exhibit extreme variance, often leading to stability issues in practice. In this work, we consider a broader class of importance weights to correct samples in off-policy learning. We propose the use of \textit{value-aware importance weights} which take into account the sample space to provide lower variance, but still unbiased, estimates under a target distribution. We derive how such weights can be computed, and detail key properties of the resulting importance weights. We then extend several reinforcement learning prediction algorithms to the off-policy setting with these weights, and evaluate them empirically.
\end{abstract}

\section{Off-policy Reinforcement Learning}
\label{sec:intro}

Value functions are central to reinforcement learning. However, most algorithms for learning a value function are limited to \textit{on-policy} learning, where a specific policy of interest must be executed in order to estimate its value function. This paradigm can be inefficient, especially in the case of long-lived agents seeking to learn about many possible ways of behaving \citep{sutton2011horde,white2015developing}. On-policy learning can also be expensive or dangerous in real-world applications such as robotics, recommendation systems \citep{maystre2023optimizing}, and healthcare \citep{liao2021off}.

In contrast, algorithms capable of \textit{off-policy} learning can learn the value function for a policy from data collected by a different policy.
This flexibility allows a lifelong learning agent to learn about many possible policies at the same time, in parallel \citep{sutton1999between,klissarov2021flexible}. It also allows an agent to learn about the optimal policy while following an exploratory policy, as in Q-learning \citep{watkins1992q}. Off-policy learning can also improve sample efficiency and stability by allowing an agent to learn from data generated by older policies, e.g., as in experience replay \citep{lin1992self,mnih2013playing,schaul2015prioritized}.
Taken to the extreme, off-policy learning can completely remove the need for environmental interaction during training, instead allowing the agent to learn offline from datasets generated by human experts, existing control algorithms, or even random behaviour \citep{levine2020offline}.

A common strategy for extending algorithms to use off-policy data is called \textit{importance sampling} \citep{kahn1950randoma,kahn1950randomb,kloek1978bayesian}.
When an action is chosen according to a different \textit{behaviour policy} than the \textit{target policy} the agent seeks to learn about, importance sampling scales the observed outcome by the ratio of the action's probability under the target and behaviour policies, known as the \textit{importance sampling ratio}.
Scaling an observed outcome (e.g., returns, policy gradients, etc.) in this way leads to an unbiased estimate of its expectation under the target policy.

However, off-policy estimators based on importance sampling tend to exhibit extreme variance in practice.
In the worst case the variance of importance sampling-based estimators can scale exponentially with the horizon of the problem \citep{liu2020understanding}.
Several approaches have been introduced to reduce variance, such as per-decision importance sampling \citep{precup2000pdis}, clipped importance sampling ratios \citep{munos2016retrace, espeholt2018impala},
self-normalized (or weighted) importance sampling \citep{rubinstein1981mc,mahmood2014weighted}, and the DICE family of algorithms that learn parametric models of importance weights \citep{nachum2019dualdice,zhang2020gendice}.
However, none of these approaches is strictly better than the others, with weighted importance sampling, clipping, and the DICE algorithms introducing bias and/or dependence on function approximation, and the variance of per-decision importance sampling can still be unacceptably large in practice \citep{precup2000pdis}.





In this work, we propose the use of \textit{value-aware importance weighting}. Like importance sampling, we correct samples by a multiplicative weight, but further take into account knowledge of the sample space (the possible outcomes of a random variable) to produce weights of lower variance. Our key contributions are as follows:
\begin{compactitem}
    \item Introduction of value-aware importance weighting as a general framework for off-policy corrections
    \item Derivation and discussion of an instance of value-aware importance weights with desirable properties
    \item Extension of existing algorithms with the derived importance weights, and empirical evaluation of the resulting algorithms.
\end{compactitem}

We emphasize our focus on correcting the update target, but not the state-visitation distribution, as it is a simpler setting where variance concerns already manifest. Nevertheless, we present preliminary results with Emphatic TD methods in Appendix \ref{app:moreexps} showcasing the generality of our estimator and its compatibility with methods which re-weight updates.

\section{Background}
\label{sec:background}


Reinforcement learning is typically modeled as a Markov decision process (MDP). At each discrete time step $t$, an agent observes the current state $S_t \in \mathcal{S}$ and selects an action $A_t \in \mathcal{A}(S_t)$, where $\mathcal{S}$ is the set of all states in the MDP, and $\mathcal{A}(s)$ is the set of available actions in state $s$. After selecting an action, the agent receives information about the next state $S_{t+1} \in \mathcal{S}$, along with a reward $R_{t+1}\in\mathbb{R}$, sampled according to the MDP's transition model $p(s',r|s,a) = P(S_{t+1}=s',R_{t+1}=r|S_t=s,A_t=a)$. Actions are selected according to a policy $\pi(a|s) = P(A_t=a|S_t=s)$, and we are interested in \textit{returns}:
\begin{equation}
G_t = \sum_{k=0}^{T-t-1}{\gamma^k R_{t+k+1}}
\label{eqn:return}
\end{equation}
where $\gamma \in [0, 1]$ and $T$ being the final time step of an episode, or $\gamma \in [0, 1)$ and $T=\infty$ in a continuing problem. Reinforcement learning control aims to compute an optimal policy $\pi^*$ which maximizes the \textit{expected} return.

\textit{Value-based methods} for reinforcement learning aim to compute value functions, which are defined to be expected returns conditioned on various quantities. Here we emphasize the \textit{action-value function}, which conditions on state $s$, immediate action $a$, and policy $\pi$ to follow thereafter:

\begin{equation}
    q_\pi(s,a) = \Ex_\pi\big[ G_t | S_t=s, A_t=a\big]
    \label{eqn:qdef}
\end{equation}

Computing value functions for a policy is denoted \textit{policy evaluation}, after which a value function may inform decision making through \textit{policy improvement}: A policy $\pi'$ derived to be greedy with respect to $q_\pi$ is in an improved policy, where $q_{\pi'}(s,a) \geq q_\pi(s,a), \forall s, a$. \textit{Policy iteration} is the process of alternating policy evaluation and policy improvement to approach an optimal policy.

Exact value function computation may be impractical due to unreasonably large state spaces and imperfect or unknown transition models. As such, solution methods typically compute approximate value functions, $Q \approx q_\pi$, informed by sampled transitions generated by interaction with the environment, e.g., with an incremental update:
\begin{equation*}
    Q(S_t,A_t) \leftarrow Q(S_t,A_t) + \alpha \big(\hat{G}_t - Q(S_t,A_t) \big)
\end{equation*}
where $\hat{G}_t$ is a sample-based estimate of the return from $(S_t, A_t)$ and following $\pi$, and $\alpha \in (0, 1]$ is the step-size.

Off-policy learning focuses on policy evaluation, but considers samples from the environment generated by a \textit{behavior policy} $\mu$, while aiming to evaluate a \textit{target policy} $\pi$. One technique to account for this discrepancy is the use of \textit{importance sampling} \citep{rubinstein1981mc,hammersley1964mc}, where a sample drawn from one distribution is re-weighted to be as if under another distribution:
\begin{align}
    &\rho_t = \frac{\pi(A_t|S_t)}{\mu(A_t|S_t)} \label{eqn:rho}\\
    &\Ex_\mu\big[\rho_t X(S_t, A_t)\big] = \Ex_\pi\big[X(S_t, A_t)\big] \nonumber
\end{align}
where $\rho_t$ is the importance weight, and we assume every action $\pi$ may select must also be possible under $\mu$. To correct the sequence of actions involved in a return, one can re-weight it with a product of these importance weights:
\begin{align*}
    &\prod_{k=t}^{T-1}{\frac{\pi(A_k|S_k)p(S_{k+1}|S_k,A_k)}{\mu(A_k|S_k)p(S_{k+1}|S_k,A_k)}} = \prod_{k=t}^{T-1}{\frac{\pi(A_k|S_k)}{\mu(A_k|S_k)}} = \prod_{k=t}^{T-1}{\rho_k} \\
    &\Ex_\mu\bigg[\bigg(\prod_{k=t}^{T-1}{\rho_k}\bigg) G_t\bigg] = \Ex_\pi\big[G_t\big] \nonumber
\end{align*}
While such re-weighted samples provide unbiased estimates under the target policy, importance weights are well known to suffer large variance \citep{sutton2018textbook}. Nevertheless, many off-policy methods for reinforcement learning are built on the idea of importance sampling \citep{precup2000pdis}, many of which propose ways to manage its large variance, often at the cost of increased bias \citep{mahmood2014weighted,munos2016retrace}.

\section{Value-aware Importance Weighting}
\label{sec:valueawareweights}


The importance weights given by Equation \ref{eqn:rho} can be seen to only depend on the behavior and target policies. This has appeal in its intuitive simplicity, as well as being generally applicable regardless of a random variable's sample space. In this section, we consider importance weights which additionally depend on a state's immediate action-values (denoted $\sparho$). We formulate it as a constrained optimization problem subject to the constraints:
\begin{enumerate}
    \item $\sum_{a}{\mu_a \sparho_a Q_a} = \sum_{a}{\pi_a Q_a}$
    \item $\sum_{a}{\mu_a \sparho_a} = 1$
\end{enumerate}
where for notational convenience, subscripts condition on a specific action, and everything is implicitly conditioned on the same (arbitrary) state. The first constraint ensures that the weighted action-values sampled under the behavior policy averages into the expected value under the target policy. The second constraint constrains the expected importance weight to be 1. This was specified because it's a property of importance sampling, and avoids arbitrary attenuation or amplification in anticipation of needing to correct a sequence. Given the variance concerns with importance sampling, a reasonable objective is to minimize $\textrm{Var}_{\mu}(\sparho)$. This choice leads to the following Lagrangian:
\begin{equation}
\mathcal{L}(\sparho, \lambda_0, \lambda_1) = \sum_{a}{\mu_a(\sparho_a - 1)^2} + \lambda_0 \Big( \sum_{a}{\pi_a Q_a} - \sum_{a}{\mu_a \sparho_a Q_a} \Big) + \lambda_1 \Big( 1 - \sum_{a}{\mu_a\sparho_a} \Big)
\label{eqn:sparholagrangian}
\end{equation}
Setting $\nabla \mathcal{L}(\sparho, \lambda_0, \lambda_1) = 0$ and solving for $\sparho$, we get:
\begin{equation}
\sparho_a = 1 + \frac{Q_a - \Ex_\mu[Q]}{\Ex_\mu[Q^2] - \Ex_\mu[Q]^2} (\Ex_\pi[Q] - \Ex_\mu[Q])
\label{eqn:sparhov}
\end{equation}
See Appendix \ref{app:derivation} for a complete derivation. We denote $\sparho$ the \textit{Sparho} estimator. Interestingly, the above expression has a form where an action-value is mean-centered and normalized by its variance (under $\mu$), and then scaled by a measure of off-policyness. We emphasize that these weights are not learned or approximated, and are computed in closed-form given a state's behavior and target policy probabilities, and its immediate action-values. Interestingly, the estimator does not directly work with the policy probabilities, and only uses expectations computed under them. This may be related to regression importance sampling \citep{hanna2018ris} in that a \textit{most likely} behavior policy (given the observed data thus far) might be implicitly used. It's unclear however if a direct connection can be made, as the above weights typically can't be expressed as a ratio of two policies.

By construction, we have:
\begin{enumerate}
    \item $\sparho = \rho$ is the unique solution to the constraints when there are 2 or fewer actions.
    \item $\sparho Q \sim \mu$ is an unbiased estimator of $\Ex_\pi[Q]$. Despite the dependence on $Q$, by satisfying the constraints it provides an immediately unbiased estimate for the current value estimates.
    \item $\textrm{Var}_{\mu}(\sparho) \leq \textrm{Var}_{\mu}(\rho)$. Ordinary importance sampling weights are in the space of solutions which satisfy the constraints, and variance is explicitly minimized.
    \item $\sparho_a = \rho_a = 1$ when $\mu = \pi$. This trivially minimizes the objective when on-policy, maintaining strict generalization of the on-policy setting akin to importance sampling.
\end{enumerate}

Despite being unbiased in estimating the expected action-value, inaccuracies in the value-function will still result in a biased estimate of the return. This is akin to the bias introduced by temporal difference methods \citep{sutton1988td}, and does not incur additional bias when correcting sampled action-values (e.g., off-policy Sarsa(0)), but will introduce bias if importance weights computed with inaccurate values are used to correct sampled rewards. Also of note, it may be possible for the importance weights to be negative. This may seem counter-intuitive in that it may lead to ``undoing" updates, but should still give the correct expectation with lower variance. Nevertheless, in our empirical evaluation we found that the weights remained all positive the majority of the time, likely due to the second constraint tending the solutions toward the positive direction. Lastly, computing $\sparho$ requires first computing $\Ex_\pi[Q]$. It may seem redundant to compute the final solution only to form a sample-based estimator for the same quantity, but we argue that the benefit lies in the need to correct a sequence, as opposed to solely an immediate action (Detailed in Section \ref{sec:multistep}). We also emphasize that the importance weights presented are for one choice of objective. We explore alternative objectives in Appendix \ref{app:altobjs}, but focus on the above estimator in this work.

\begin{figure}[ht]
    \centering
    \begin{subfigure}[b]{0.35\textwidth}
        \centering
        \includegraphics[width=\textwidth]{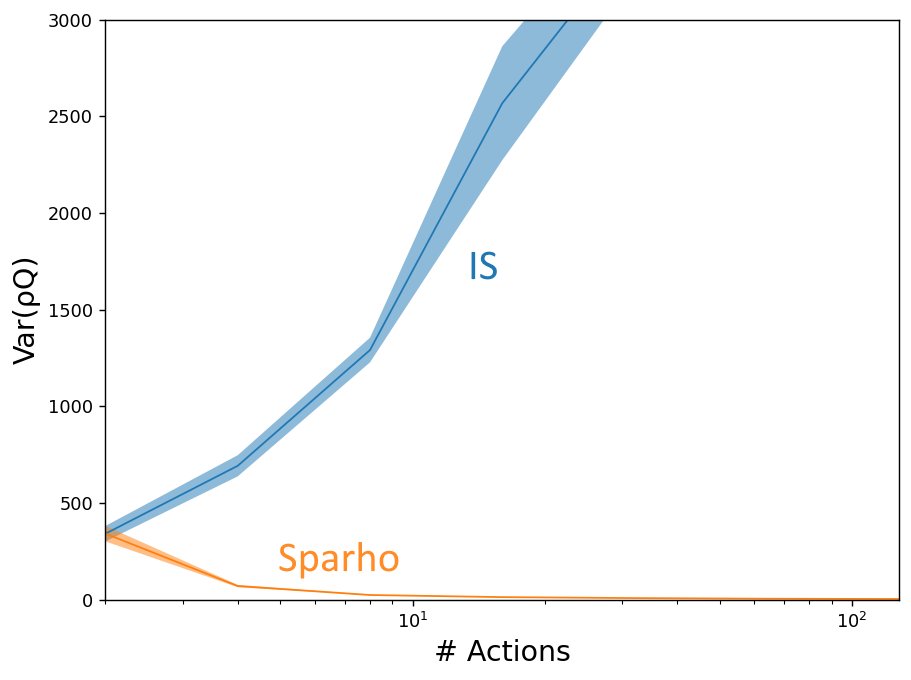}
    \end{subfigure}
    \begin{subfigure}[b]{0.35\textwidth}
        \centering
        \includegraphics[width=\textwidth]{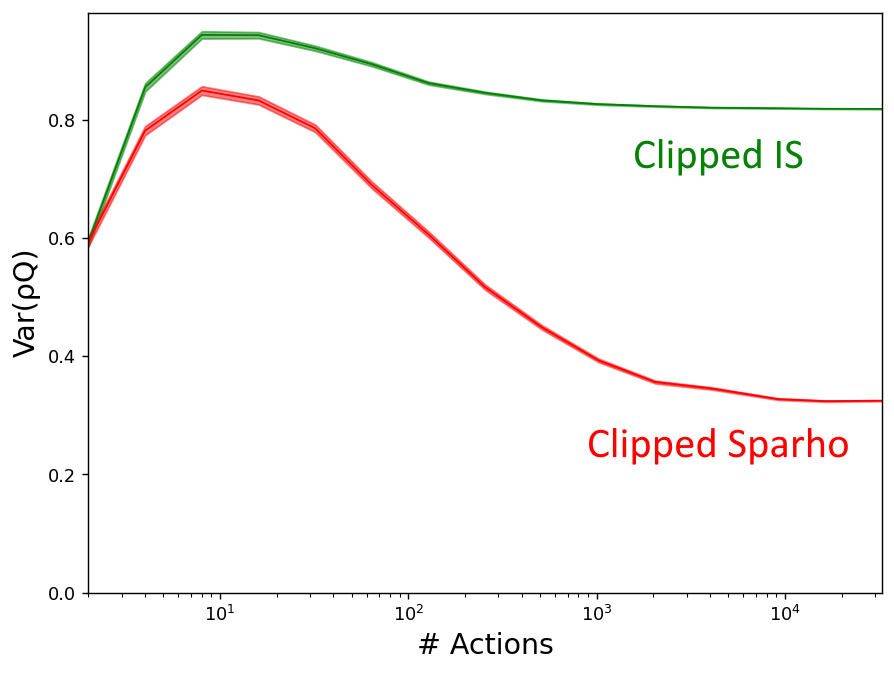}
    \end{subfigure} \\
    \begin{subfigure}[b]{0.35\textwidth}
        \centering
        \includegraphics[width=\textwidth]{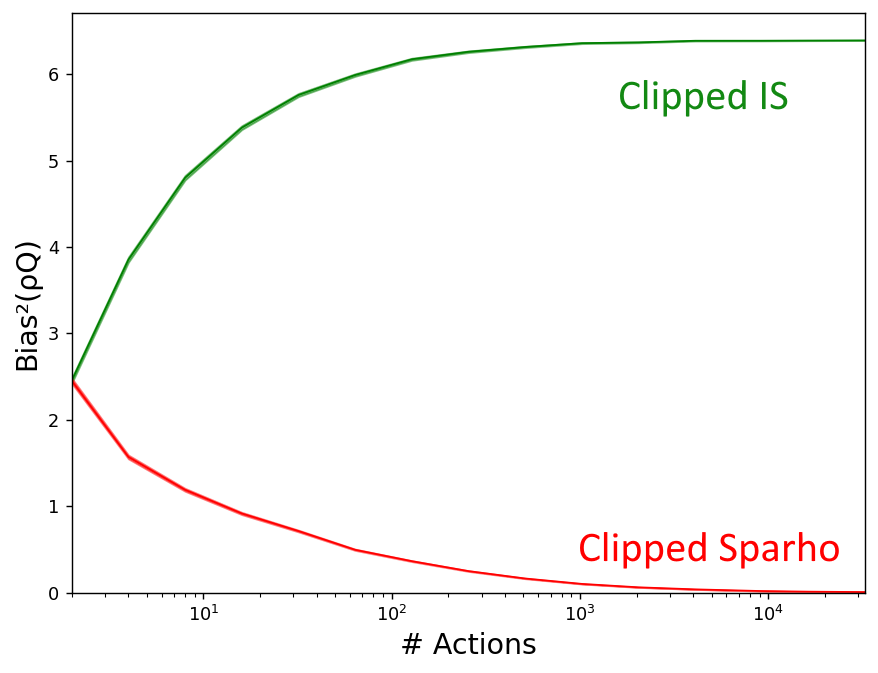}
    \end{subfigure}
    \begin{subfigure}[b]{0.35\textwidth}
        \centering
        \includegraphics[width=\textwidth]{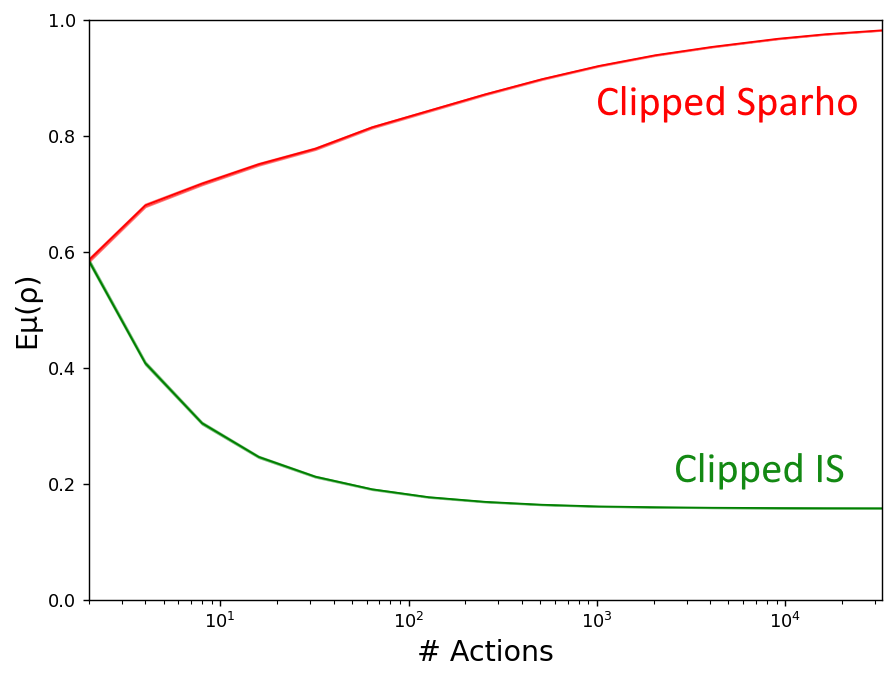}
    \end{subfigure}
    \caption{Various statistics against size of the action space, measured across randomly generated bandit instances. Results are averaged over 10,000 instances, and shaded regions represent one standard error.}
    \label{fig:bandit_results}
\end{figure}

Here we empirically evaluate the above estimator in an off-policy bandit prediction setting to gain intuition on the potential variance reduction, and observe how it might scale with the size of the action space. Bandit instances are randomly generated by sampling $\mu$ and $\pi$ according to $\textrm{softmax}(\mathcal{N}_{|\mathcal{A}|\times 1}(0, \beta))$, and $Q$ according to $\beta + \mathcal{N}_{|\mathcal{A}|\times 1}(0, \beta))$, with $\beta$ tuning the determinism of the policies and the spread of the action-values. Using $\beta = 2$, we swept over action space sizes from 2 to 32,768. For each action space, we generated 10,000 bandit instances and measured $\textrm{Var}(\rho Q)$ with $\rho$ as the importance sampling and Sparho estimators. We additionally compare clipped versions of the above estimator, akin to Retrace and Vtrace estimators \citep{munos2016retrace, espeholt2018impala} For the clipped versions of the previous estimators, we additionally measure $\textrm{Bias}^2(\rho Q)$ and $\Ex_\mu[\rho]$. Results are shown in Figure \ref{fig:bandit_results}.

It can be seen that the Sparho estimator has substantially lower variance than the importance sampling estimator. Perhaps surprisingly, we observe a \textit{decrease} in variance with the size of the action space, despite the exponential growth in that of the importance sampling estimator. With the clipped estimators, we see a substantial decrease in variance. However, the clipped Sparho estimator had lower variance, substantially lower bias, and maintained a larger expected importance weight than the clipped importance sampling estimator. Of note, the clipped Sparho estimator similarly exhibited very favorable scaling with the size of the action space.

The above results assume access to the true action-values, computing relevant statistics in closed form for a random bandit instance. We note that results are similar with learned values, and show a representative example in Appendix \ref{app:moreexps}.

\section{Multi-step corrections}
\label{sec:multistep}

The action-value definition can be expressed in terms of successor values, denoted the \textit{Bellman equation for $q_\pi$}:
\begin{equation}
q_\pi(s, a) = \sum_{s', r}{p(s',r|s,a) \Big( r + \gamma \sum_{a'}{\pi(a'|s')q_\pi(s', a')} \Big)}
\label{eqn:qbellman}
\end{equation}

We have that $\sum_{a}{\pi(a|s)q_\pi(s, a)} = \sum_{a}{\mu(a|s)\sparho(s, a)q_\pi(s, a)}$, giving:
\begin{equation}
q_\pi(s, a) = \sum_{s', r}{p(s',r|s,a) \Big( r + \gamma \sum_{a'}{\mu(a'|s')\sparho(s', a')q_\pi(s', a')} \Big)}
\label{eqn:sparhobellman}
\end{equation}

Should we expand the recursion multiple times, and then draw samples according to behavior policy $\mu$, we get a multi-step estimator analogous to per-decision importance sampling:
\begin{align}
G_t &= R_{t+1} + \gamma \sparho_{t+1} G_{t+1} \nonumber \\
G_t &= R_{t+1} + \gamma \sparho_{t+1} R_{t+2} + \gamma^2 \sparho_{t+1} \sparho_{t+2} R_{t+3} + \gamma^3 \sparho_{t+1} \sparho_{t+2} \sparho_{t+3} R_{t+4} + \dots
\label{eqn:pdsparho}
\end{align}

Such a form, like importance sampling, conveniently lends itself to online, incremental algorithms (e.g., TD($\lambda$)). An important note, however, is that $\sparho_{k}$ is computed based on the current value estimates for state $S_{k}$, and if values are inaccurate, the expansion of the Bellman operator may result in a different quantity than what was used to compute the importance weights. This suggests that additional bias may be introduced in this multi-step extension, but also that the true value function remains the fixed-point. Of note, it's unclear whether the resulting operator is still a contraction, but we have not come across any example of multiple fixed-points in our empirical evaluation.

In the one-step case, a sample of Equation \ref{eqn:sparhobellman} gives Expected Sarsa, which is convergent under standard assumptions. The multi-step case is a bit more involved due to the form of the Sparho estimator, but in a loop-free episodic MDP, we have that a state-action pair preceding a terminal state will be convergent as it only predicts the immediate reward. By induction, we get a cascade of one-step Expected Sarsa estimators with eventual stable bootstrap targets. It is uncertain whether we have such guarantee for general MDPs due to the relatively complex expression for the importance weights, the existence of negative weights, and the non-stationarity of the weights as $Q$ changes.

\begin{figure}[ht]
    \centering
    \includegraphics[width=0.5\textwidth]{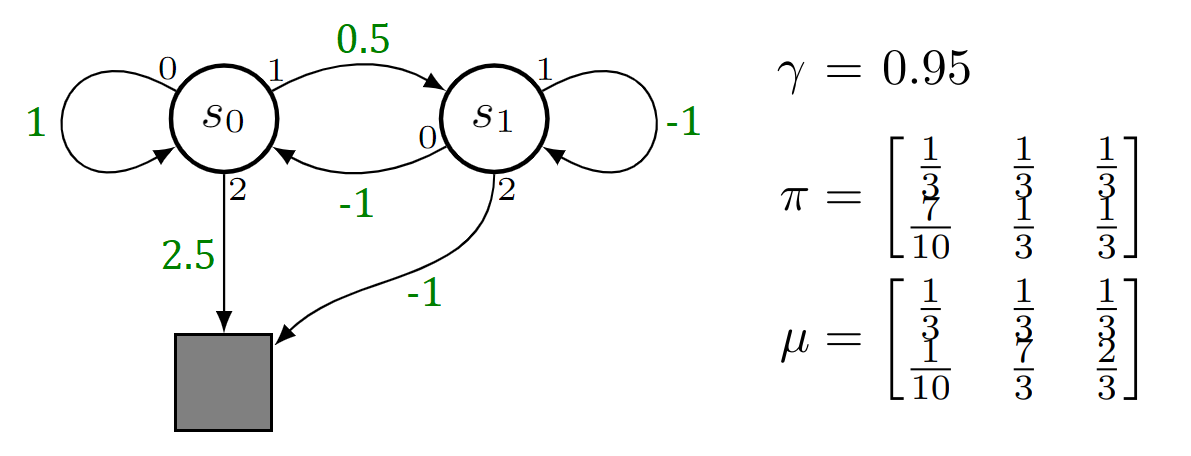}
    \caption{Two-state MDP for visualizing expected multi-step updates.}
    \label{fig:2statemdp}
\end{figure}

\begin{figure}[ht]
    \centering
    \begin{subfigure}[b]{0.33\textwidth}
        \centering
        \includegraphics[width=\textwidth]{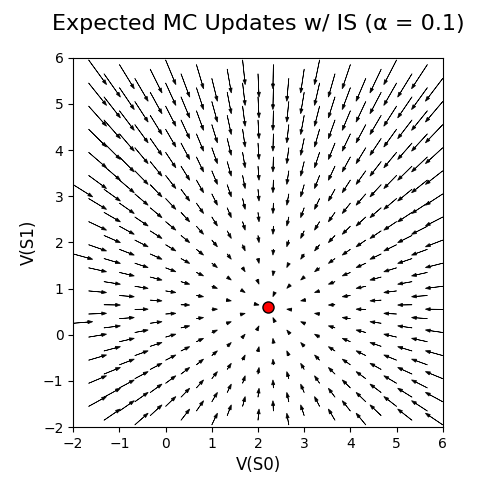}
    \end{subfigure}
    \hfill
    \begin{subfigure}[b]{0.33\textwidth}
        \centering
        \includegraphics[width=\textwidth]{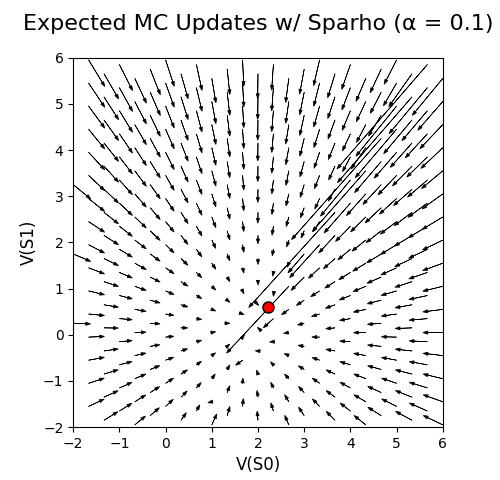}
    \end{subfigure}
    \hfill
    \begin{subfigure}[b]{0.33\textwidth}
        \centering
        \includegraphics[width=\textwidth]{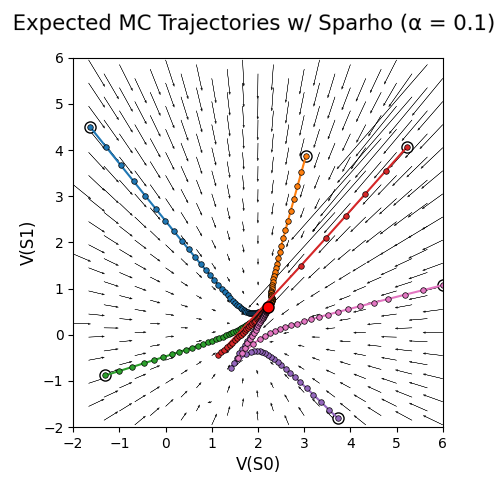}
    \end{subfigure}
    \caption{Expected every-visit Monte Carlo updates for the per-decision importance sampling and Sparho estimators. The red circle denotes the fixed-point $v_\pi$, and sample trajectories of the Sparho estimator are visualized on the right.}
    \label{fig:mc_odes}
\end{figure}

We compute the \textit{expected} every-visit Monte Carlo update for the per-decision Sparho estimator given by Equation \ref{eqn:pdsparho} from various starting values in a two-state MDP. This lets us visualize update directions and gain intuition for the bias introduced in the Monte Carlo extreme of the multi-step case. We use a step size $\alpha=0.1$, and show the importance sampling estimator for reference. Figure \ref{fig:2statemdp} precisely specifies the MDP, and Figure \ref{fig:mc_odes} visualizes the expected updates.

As one might expect, the unbiased importance sampling estimator points directly at the fixed-point, and the arrows would directly jump there if $\alpha = 1$. The multi-step Sparho estimator generally has similar magnitude updates, but points in slightly different directions relating to its bias. Visualizing trajectories, the values seem to rush to an area near the fixed-point, before turning to approach it. Such a path may be an artifact of projecting higher-dimensional action-values into two state-values for visualization. Of particular note is the top right corner, where large jumps occur despite the small step size. We suspect that such large magnitude updates are due to interference between the transition matrix $P_\mu$ and the importance weights in the matrix $(I - \gamma P_\mu \textrm{diag}(\sparho))^{-1}$. Under importance sampling, we have that $P_\pi = P_\mu \textrm{diag}(\rho)$, but not with $\sparho$. It's difficult to comment on the generality and extent of the phenomenon, as $\sparho$ keeps changing, but in stochastic empirical evaluation (See Section \ref{sec:experiments}) such jumps appeared indistinguishable from typical return variance, and did not appear to cause divergence.

\begin{figure}[ht]
    \centering
    \begin{subfigure}[b]{0.33\textwidth}
        \centering
        \includegraphics[width=\textwidth]{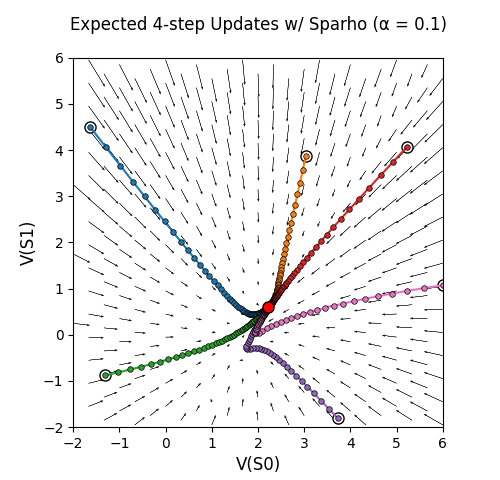}
    \end{subfigure}
    \hfill
    \begin{subfigure}[b]{0.33\textwidth}
        \centering
        \includegraphics[width=\textwidth]{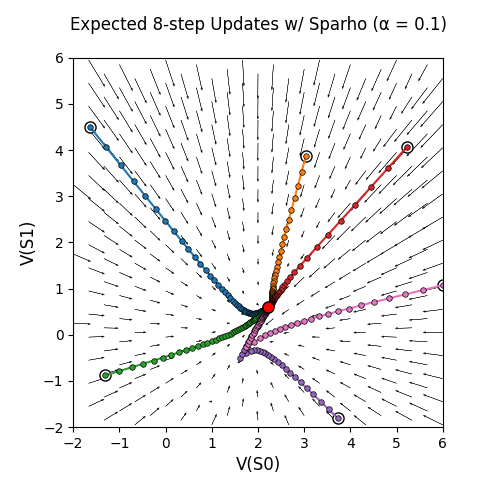}
    \end{subfigure}
    \hfill
    \begin{subfigure}[b]{0.33\textwidth}
        \centering
        \includegraphics[width=\textwidth]{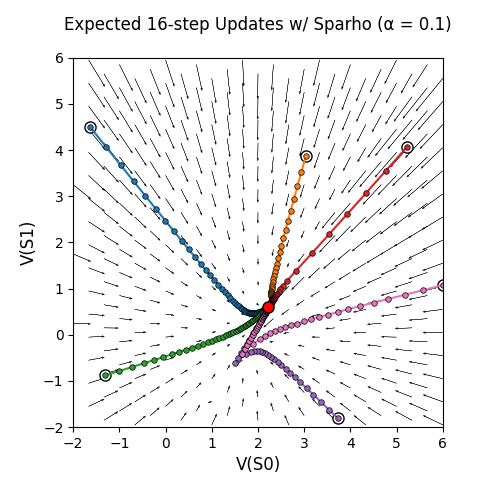}
    \end{subfigure}
    \caption{Expected $n$-step TD updates for the per-decision Sparho estimator. The red circle denotes the fixed-point $v_\pi$, with sample trajectories from a variety of starting points are visualized.}
    \label{fig:nstep_odes}
\end{figure}

Of note, $n$-step temporal difference (TD) estimators can attenuate the momentary large magnitude updates:
\begin{align}
G_{t:h} &= R_{t+1} + \gamma \sparho_{t+1} G_{t+1:h} \nonumber \\
G_{t:t} &= Q(S_t, A_t)
\label{eqn:nsteppdsparho}
\end{align}
where $h = t + n$. Such attenuation is expected as it interpolates Monte Carlo returns with the one-step extreme, which was previously acknowledged to be Expected Sarsa. Figure \ref{fig:nstep_odes} explores this for $n \in \{4, 8, 16\}$ for the above MDP. Perhaps surprisingly, the 16-step TD update has no noticeable large jumps, and appears closer to the 4-step updates than Monte Carlo despite the MDP having an expected trajectory length of around 2 steps. The top-right sample trajectory required $n$ close to 100 to start overshooting the fixed-point akin to the Monte Carlo visualization.

\section{Comparison with Online Off-policy Methods}
\label{sec:experiments}

In this section, we empirically compare the multi-step Sparho estimator with importance sampling in a variety of environments. We specifically focus on the online, incremental setting with eligibility traces \citep{sutton1988td, precup2000pdis}, starting with the following $\lambda$-return:
\begin{equation}
    G_t^{\lambda} = R_{t+1} + \gamma \Big( \lambda \big( \rho_{t+1} G_{t+1}^{\lambda} + \Ex_\pi [Q(S_{t+1}, \cdot)] - \rho_{t+1} Q(S_{t+1}, A_{t+1})\big) + (1 - \lambda) \Ex_\pi [Q(S_{t+1}, \cdot)] \Big)
    \label{eqn:lambdareturndef}
\end{equation}
which can be re-written as the following sum of one-step Expected Sarsa TD errors, weighted by an eligibility trace:
\begin{equation}
    G_t^{\lambda} = Q(S_t, A_t) + \sum_{k=t}^{\infty}{\big(R_{k+1} + \gamma \Ex_\pi [Q(S_{k+1}, \cdot)] - Q(S_k, A_k) \big) \prod_{i=t+1}^{k}{\gamma \lambda \rho_i}}
    \label{eqn:lambdareturntderrs}
\end{equation}
We denote the incremental algorithm using the above $\lambda$-return as Q($\lambda$). It corresponds with the $n$-step per-decision estimator given by Equation \ref{eqn:pdsparho}, but with $\Ex_\pi [Q(S_{t+1}, \cdot)] - \rho_{t+1} Q(S_{t+1}, A_{t+1})$ appended to every reward beyond the immediate one. The appended terms have zero expected value and can be viewed as control variates, leading to a doubly-robust estimator \citep{jiang2016doublyrobust, thomas2016magic, deasis2018pdcv}. They're central to the presence of one-step Expected Sarsa TD errors in Equation \ref{eqn:lambdareturntderrs}. Without such terms, we would get Sarsa($\lambda$) with per-decision importance sampling \citep{precup2000pdis}, where $\Ex_\pi [Q(S_{k+1}, \cdot)] \rightarrow \rho_{k+1} Q(S_{k+1}, A_{k+1})$, leading to additional variance on the order of what is presented in Figure \ref{fig:bandit_results} 
This choice of $\lambda$-return is in anticipation of an analogous Sparho estimator, which requires computing the expected action-value under the target policy anyways, leading to a fairer comparison. We denote the Sparho variant of the above estimator as Sparho($\lambda$):
\begin{equation}
    G_t^{\lambda} = Q(S_t, A_t) + \sum_{k=t}^{\infty}{\big(R_{k+1} + \gamma \Ex_\pi [Q(S_{k+1}, \cdot)] - Q(S_k, A_k) \big) \prod_{i=t+1}^{k}{\gamma \lambda \sparho_i}}
    \label{eqn:sparholambda}
\end{equation}
Additionally, we compare against clipped variants of each estimator, where the importance weights in the above $\lambda$-returns are clipped to the range $[0, 1]$. Clipping them in Equation \ref{eqn:lambdareturntderrs} gives Retrace($\lambda$) \citep{munos2016retrace}, and we denote the clipped variant of Equation \ref{eqn:sparholambda} as ReSparho($\lambda$).

\begin{figure}[ht]
    \centering
    \begin{subfigure}[b]{0.33\textwidth}
        \centering
        \includegraphics[width=\textwidth]{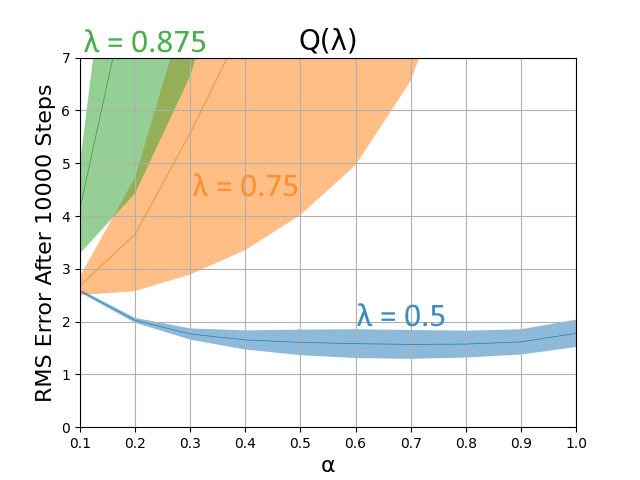}
        \caption{Q($\lambda$) with 8 actions}
    \end{subfigure}
    \hfill
    \begin{subfigure}[b]{0.33\textwidth}
        \centering
        \includegraphics[width=\textwidth]{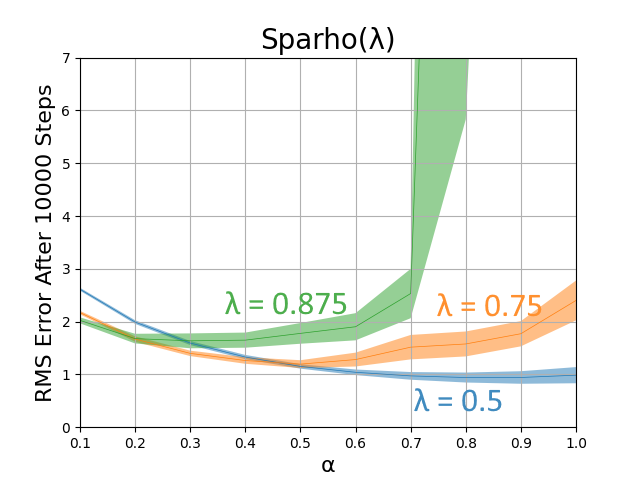}
        \caption{Sparho($\lambda$) with 8 actions}
    \end{subfigure}
    \hfill
    \begin{subfigure}[b]{0.33\textwidth}
        \centering
        \includegraphics[width=\textwidth]{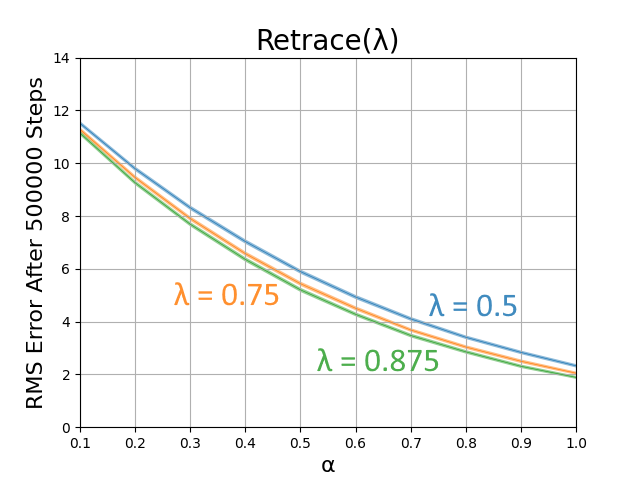}
        \caption{Retrace($\lambda$) with 32 actions}
    \end{subfigure}
    \begin{subfigure}[b]{0.33\textwidth}
        \centering
        \includegraphics[width=\textwidth]{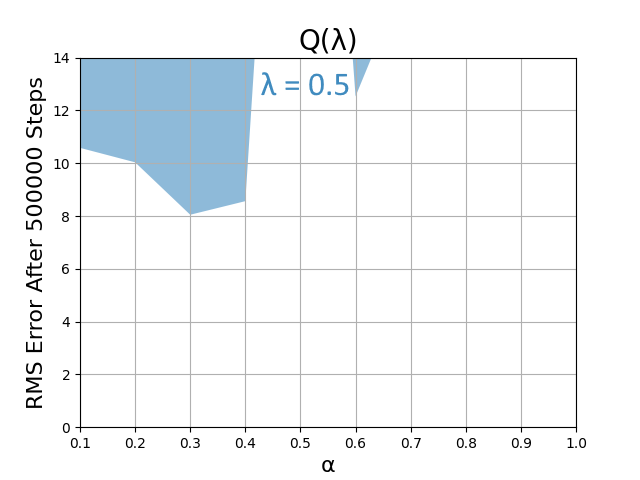}
        \caption{Q($\lambda$) with 32 actions}
    \end{subfigure}
    \hfill
    \begin{subfigure}[b]{0.33\textwidth}
        \centering
        \includegraphics[width=\textwidth]{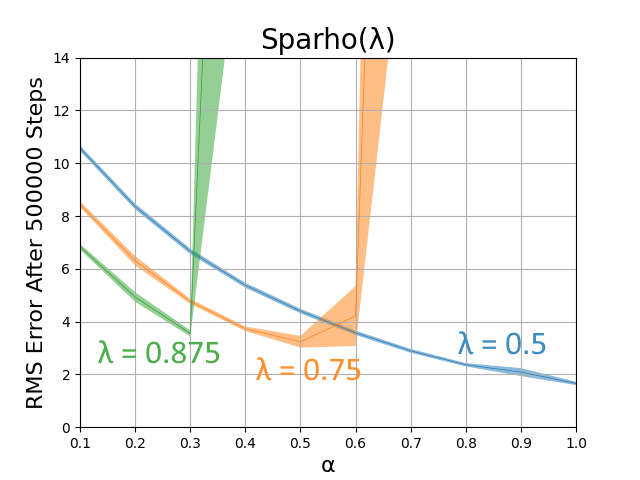}
        \caption{Sparho($\lambda$) with 32 actions}
    \end{subfigure}
    \hfill
    \begin{subfigure}[b]{0.33\textwidth}
        \centering
        \includegraphics[width=\textwidth]{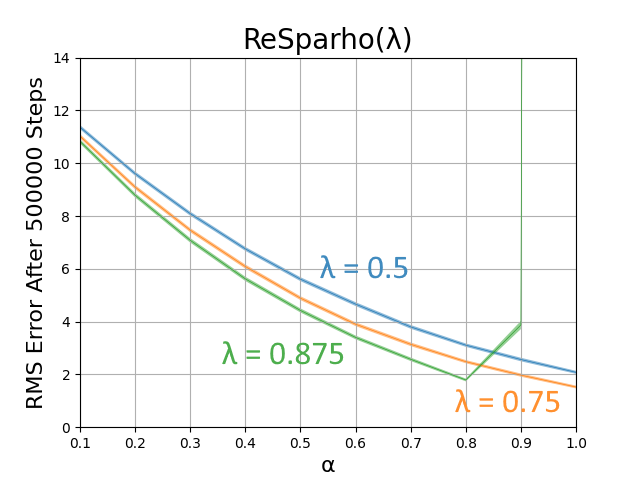}
        \caption{ReSparho($\lambda$) with 32 actions}
    \end{subfigure}
    \caption{Results of an ablation study in the Path World environment for $|\mathcal{A}|\in\{8, 32\}$. Results are averaged over 30 runs, and shaded regions represent one standard error.}
    \label{fig:pw_results}
\end{figure}

First, we evaluate the aforementioned algorithms in \textit{Path World}. The environment can be thought of as a fully-connected feed-forward neural network with one input and one output. Each neuron is a state and each connection is an action, with the input being the initial state and the output the terminal state. The width of the network represents the size of the action space, $|\mathcal{A}|$, and the depth gives the number of decisions per episode. Each transition is assigned a fixed reward, and every run, behavior and target policy probabilities are generated per-state with the procedure of Section \ref{sec:valueawareweights}'s bandit experiments with $\beta = 1$. We fix the depth to 5, consider $|\mathcal{A}| \in \{8, 32\}$, and performed a systematic sweep over step size $\alpha$ and trace-decay rate $\lambda$. The root-mean-square (RMS) error between the learned values an the true values was measured after 10,000 steps for $|\mathcal{A}|=8$, and after 500,000 steps for $|\mathcal{A}|=32$. 30 independent runs were performed and results can be seen in Figure \ref{fig:pw_results}.

It can be seen that Q($\lambda$) suffers variance issues for $\lambda > 0.5$, and struggles further as the action space grows. In contrast, Sparho($\lambda$)'s reduced variance enables the use of larger $\lambda$, but may be unstable for large $(\alpha, \lambda)$ pairs. For the clipped variants, we see substantially reduced variance, but a slower rate of learning due to the additional bias introduced by an expected importance weight $< 1$. Of note, ReSparho($\lambda$) still outperforms Retrace($\lambda$) across most parameter settings, consistent with the bandit results demonstrating a larger expected importance weight, but exhibits similar instability should the $(\alpha, \lambda)$ pair get too large. Such instability with large $(\alpha, \lambda)$ pairs are likely due to the potential momentary large-magnitude updates observed in Figure \ref{fig:mc_odes}, from the interference between $P_\mu$ and $\textrm{diag}(\sparho)$.

\begin{figure}[ht]
    \begin{subfigure}[b]{0.49\textwidth}
        \centering
        \includegraphics[width=\textwidth]{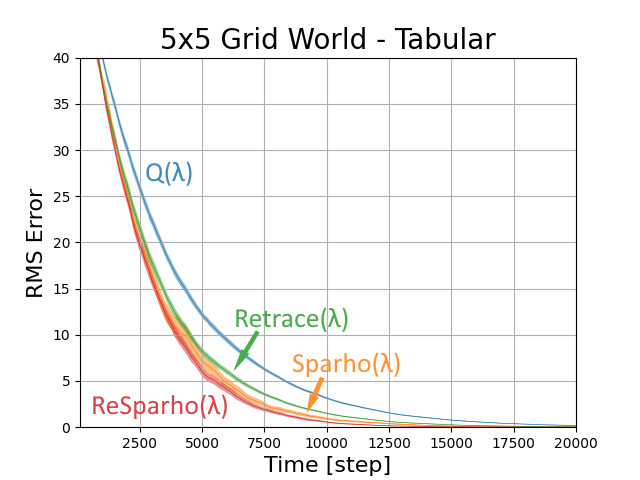}
        \caption{Tabular 5$\times$5 Grid World results}
        \label{fig:gw4_results}
    \end{subfigure}
    \hfill
    \begin{subfigure}[b]{0.49\textwidth}
        \centering
        \includegraphics[width=\textwidth]{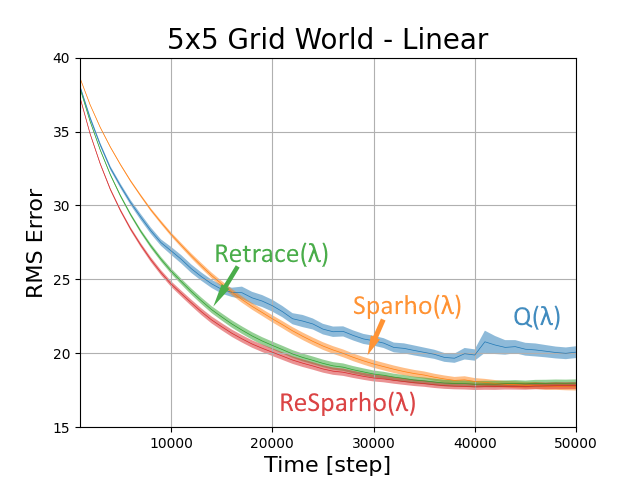}
        \caption{Linear 5$\times$5 Grid World results}
        \label{fig:gw8_results}
    \end{subfigure}
    \caption{Results in the 5 $\times$ 5 Grid World environment with tabular and linear representations. Results are averaged over 100 runs, and shaded regions represent one standard error.}
\end{figure}

Next, we consider a 5 $\times$ 5 grid world with the center as the start state and two opposite corners as terminal states. Moving off of the grid keeps the agent in place, and a reward of -1 is received at each step. We first use a tabular representation with 4-directional movement, and we consider behavior and target policies which favor a specified action with probability $1 - \epsilon$, and behave uniformly random otherwise. In contrast with the Path World environment, this MDP sees considerably many state revisitations within an episode. Performing a systematic sweep over step size and trace-decay rate, we present learning curves for the best parameter setting in terms of RMS error after 20,000 steps. Results are averaged across 100 runs, and can be seen in Figure \ref{fig:gw4_results}.

To see how results scale to more difficult problems, we use a variant of the environment with 8-directional movement, and use feature vectors for use with linear function approximation. For each run, a random 16-bit binary feature vector is generated per state consisting of exactly 8 ones. Such a choice prevents perfectly representing the true values, and allows us to see the extent of potential bias the multi-step Sparho operator may introduce in the linear fixed-point. A similar parameter sweep was performed, and learning curves are presented for the best parameter setting in terms of RMS error after 50,000 steps. Results are averaged across 100 runs, and can be seen in Figure \ref{fig:gw8_results}.

With the tabular representation, we see that Q($\lambda$) was slowest to converge. Retrace($\lambda$) substantially improves over it, with both Sparho($\lambda$) and ReSparho($\lambda$) performing even better. Of note, improved performance here largely comes from the ability to tolerate larger trace-decay rates without becoming unstable, and explains why the curves for the Sparho estimators may appear to have higher variance. In the case of the clipped estimators, assuming $\lambda \leq 1$, the rate of convergence is bottle-necked by the expected importance weight magnitude. With the linear representation, we see a similar trend, but with Retrace($\lambda$) learning quicker than Sparho($\lambda$). Of note, despite the aformentioned bias introduced by the multi-step Sparho operator, there is no noticeable impact on the final RMS error achieved, with every algorithm but Q($\lambda$) converging to a similar distance to the true value function. This is likely due to the expectation-correction behavior \citep{harutyunyan2016newQ} of the control variates in Equation \ref{eqn:lambdareturndef}.

\begin{figure}[ht]
    \centering
    \begin{subfigure}[b]{0.35\textwidth}
        \centering
        \includegraphics[width=\textwidth]{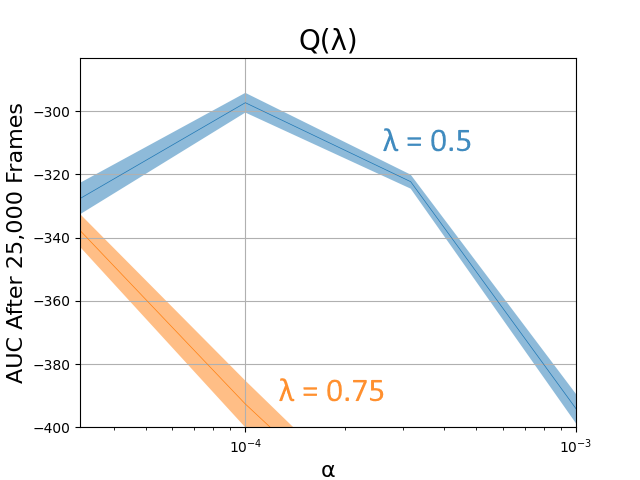}
    \end{subfigure}
    \begin{subfigure}[b]{0.35\textwidth}
        \centering
        \includegraphics[width=\textwidth]{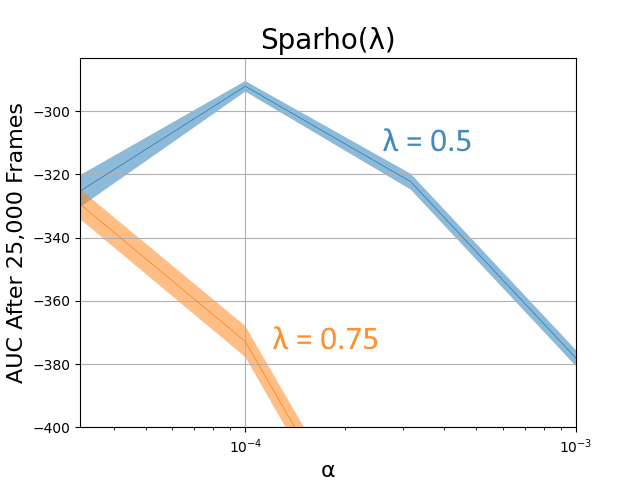}
    \end{subfigure} \\
    \begin{subfigure}[b]{0.35\textwidth}
        \centering
        \includegraphics[width=\textwidth]{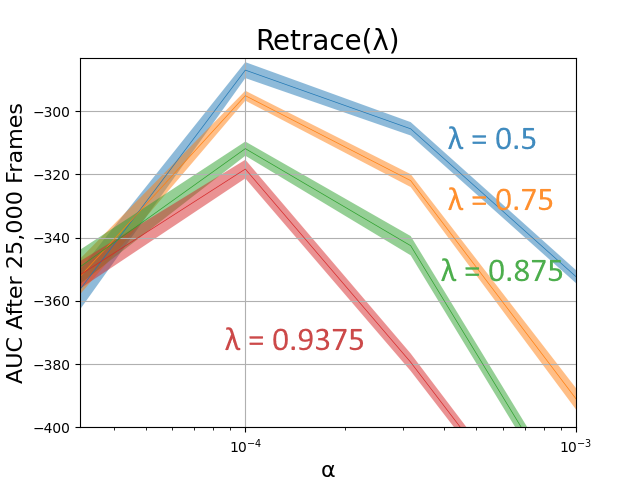}
    \end{subfigure}
    \begin{subfigure}[b]{0.35\textwidth}
        \centering
        \includegraphics[width=\textwidth]{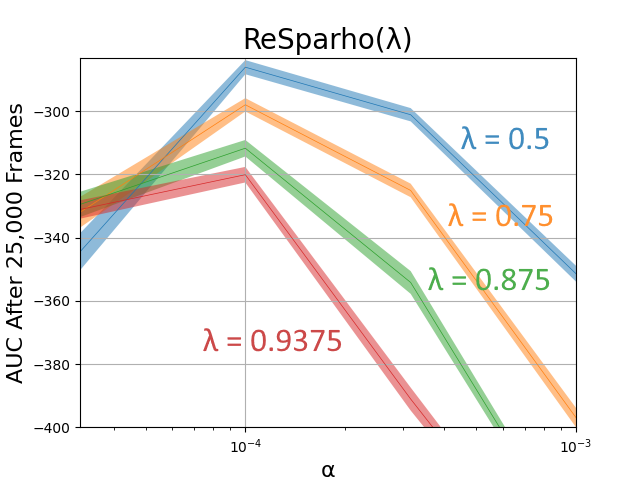}
    \end{subfigure}
    \caption{Results in the Acrobot-v1 environment with a hidden layer width of 512. Results are averaged over 30 runs, and shaded regions represent one standard error.}
    \label{fig:acrobot_results}
\end{figure}

Lastly, we evaluate the algorithms in the \textit{Acrobot-v1} environment \citep{brockman2016gym}, to demonstrate whether previous results extend toward control problems with non-linear function approximation. We used an $\epsilon$-greedy behavior policy where $\epsilon$ was annealed from 1 to 0.1 over 25,000 steps, with a target policy deterministically greedy with respect to the current value estimates. Of note, all of the above algorithms are identical in the on-policy case. As a result, we evaluate each algorithm based on the the first 25,000 steps, where there is a larger degree of off-policyness. Specifically, for each step we record the mean of the last 10 episodes, and measure the area under the curve (AUC). A 2-hidden-layer fully-connected neural network with \textit{tanh} activations was used to represent the action-values, and we performed a sweep over step size, trace-decay rate, and hidden layer width. Averaged across 30 runs, the results for the best hidden layer width can be seen in Figure \ref{fig:acrobot_results}.

It can be seen that Sparho($\lambda$) outperforms Q($\lambda$), with lower variance in the episodic returns. ReSparho($\lambda$) performs similar to Retrace($\lambda$) \textit{when properly tuned}, but exhibits flatter sensitivity curves in the direction of smaller $\alpha$. Similarities in performance are likely due to relatively small degrees of off-policyness ($\epsilon$-greedy vs. greedy), the environment only having three actions, and how control performance might not require accurate value estimates. Nevertheless, all of the empirical evaluation taken together suggests that the improvements here are likely due to better value estimation.

\section{Conclusion and Future Work}

In this work, we investigated the use of value-aware importance weights as a lower variance alternative to importance sampling. We derived a minimum-variance instance of such weights, empirically characterized the potential variance reduction, and analyzed the behavior of its per-decision multi-step extension. We then evaluated the online learning capabilities of our novel estimator and demonstrated improvements in both off-policy prediction and control, where the difference between pairs of algorithms is the substitution of value-aware importance weights.

This opens several avenues for future work. Our work focused on one instantiation of value-aware importance weights, but there are many alternate objectives one can consider in deriving new estimators, e.g., a non-negative constraint, adapted mixtures of importance sampling and Sparho for higher ($\alpha$, $\lambda$) pairs, etc. Much of our analysis was through empirical evaluation, leaving room for theoretical analysis of the Sparho estimator and other comparable importance weights. We further acknowledge complimentary similarities with the more robust doubly-robust estimator \citep{farajtabar2018mrdr}, where one instead searches the space of value estimates to minimize the estimator's variance. This suggests the possibility of searching through the combined spaces of importance weights and value-estimates to further reduce variance. Lastly, it's worth exploring other extensions for the multi-step case. Instead of incurring bias from only considering immediate action-values, a trajectory-aware algorithm could use future sampled information in place of the action-values of selected actions. However, it's unclear if such a procedure can produce an online algorithm.

\subsection*{Acknowledgements}

The authors were generously supported by Amii, NSERC, and CIFAR, and would like to thank Arsalan Sharifnassab and Huizhen Yu for insights and discussions contributing to the results presented in this paper. The authors further thank the reviewers for valuable feedback during the review process.

\bibliography{collas2023_conference}
\bibliographystyle{collas2023_conference}

\newpage
\appendix
\section{Sparho Derivation}
\label{app:derivation}
We begin with the Lagrangian from Equation \ref{eqn:sparholagrangian}.
\begin{equation*}
\mathcal{L}(\sparho, \lambda_0, \lambda_1) = \sum_{a}{\mu_a(\sparho_a - 1)^2} + \lambda_0 \Big( \sum_{a}{\pi_a Q_a} - \sum_{a}{\mu_a \sparho_a Q_a} \Big) + \lambda_1 \Big( 1 - \sum_{a}{\mu_a\sparho_a} \Big)
\end{equation*}
We can rewrite this in matrix form as follows:
\begin{equation}
\mathcal{L}(\sparhov, \lambda_0, \lambda_1) = (\sparhov - 1)^T D_{\mu} (\sparhov - 1) + \lambda_0 \big( \piv^T \Qv - (D_\mu \Qv)^T \sparhov \big) + \lambda_1 (1 - (D_\mu \onev)^T \sparhov)
\label{eqn:sparholagrangianmat}
\end{equation}
where $D_\mu$ is a diagonal matrix with entries of $\muv$ along its diagonal. We assume $\muv$ is soft, i.e., has a non-zero probability for each action, ensuring $D_\mu$'s inverse exists. Setting $\nabla \mathcal{L}(\sparhov, \lambda_0, \lambda_1) = 0$ and solving for $\sparhov$, we get:
\begin{align}
\nonumber
\nabla \mathcal{L}(\sparhov, \lambda_0, \lambda_1) = 0 &= 2 D_\mu \sparhov - 2 \muv - D_\mu \Qv \lambda_0 - D_\mu \onev \lambda_1 \\
\nonumber
0 &= D_\mu \sparhov - \muv - D_\mu \Qv \frac{\lambda_0}{2} - D_\mu \onev \frac{\lambda_1}{2} \\
\nonumber
0 &= D_\mu^{-1} D_\mu \sparhov - D_\mu^{-1} \muv - D_\mu^{-1} D_\mu \Qv \frac{\lambda_0}{2} - D_\mu^{-1} D_\mu \onev \frac{\lambda_1}{2} \\
\nonumber
0 &= \sparhov - \onev - \Qv \frac{\lambda_0}{2} - \onev \frac{\lambda_1}{2} \\
\sparhov &= \onev + \Qv \frac{\lambda_0}{2} + \onev \frac{\lambda_1}{2}
\label{eqn:sparhowithmultipliers}
\end{align}
Multiply both sides of Equation \ref{eqn:sparhowithmultipliers} by $(D_\mu \Qv)^T$ and apply one of the constraints:
\begin{align}
\nonumber
(D_\mu \Qv)^T \sparhov &= (D_\mu \Qv)^T \onev + (D_\mu \Qv)^T \Qv \frac{\lambda_0}{2} + (D_\mu \Qv)^T \onev \frac{\lambda_1}{2} \\
\nonumber
(D_\mu \Qv)^T \sparhov - (D_\mu \Qv)^T \onev &= (D_\mu \Qv)^T \Qv \frac{\lambda_0}{2} + (D_\mu \Qv)^T \onev \frac{\lambda_1}{2} \\
\piv^T \Qv - \muv^T \Qv &= \muv^T (\Qv \circ \Qv) \frac{\lambda_0}{2} + \muv^T \Qv \frac{\lambda_1}{2}
\label{eqn:sparhowithmultipliers_step1}
\end{align}
Multiply both sides of Equation \ref{eqn:sparhowithmultipliers} by $(D_\mu \onev)^T$ and apply the other constraint:
\begin{align}
\nonumber
(D_\mu \onev)^T \sparhov &= (D_\mu \onev)^T \onev + (D_\mu \onev)^T \Qv \frac{\lambda_0}{2} + (D_\mu \onev)^T \onev \frac{\lambda_1}{2} \\
\nonumber
(D_\mu \onev)^T \sparhov - (D_\mu \onev)^T \onev &= (D_\mu \onev)^T \Qv \frac{\lambda_0}{2} + (D_\mu \onev)^T \onev \frac{\lambda_1}{2} \\
0 &= \muv^T \Qv \frac{\lambda_0}{2} + \frac{\lambda_1}{2}
\label{eqn:sparhowithmultipliers_step2}
\end{align}
We can now use Equations \ref{eqn:sparhowithmultipliers_step1} and \ref{eqn:sparhowithmultipliers_step2} to solve for the Lagrange multipliers:
\begin{align}
\nonumber
\begin{bmatrix}
\muv^T (\Qv \circ \Qv) & \muv^T \Qv \\
\muv^T \Qv & 1
\end{bmatrix}
\begin{bmatrix}
\frac{\lambda_0}{2} \\
\frac{\lambda_1}{2}
\end{bmatrix} &= 
\begin{bmatrix}
\piv^T \Qv - \muv^T \Qv \\
0 \\
\end{bmatrix} \\
\nonumber
\begin{bmatrix}
\frac{\lambda_0}{2} \\
\frac{\lambda_1}{2}
\end{bmatrix} &=
\begin{bmatrix}
\muv^T (\Qv \circ \Qv) & \muv^T \Qv \\
\muv^T \Qv & 1
\end{bmatrix}^{-1}
\begin{bmatrix}
\piv^T \Qv - \muv^T \Qv \\
0 \\
\end{bmatrix} \\
\frac{\lambda_0}{2} &= \frac{\piv^T \Qv - \muv^T \Qv}{\muv^T (\Qv \circ \Qv) - (\muv^T \Qv)^2}
\label{eqn:multiplier0} \\
\frac{\lambda_1}{2} &= -\muv^T \Qv \frac{\piv^T \Qv - \muv^T \Qv}{\muv^T (\Qv \circ \Qv) - (\muv^T \Qv)^2}
\label{eqn:multiplier1}
\end{align}
Substituting Equations \ref{eqn:multiplier0} and \ref{eqn:multiplier1} back into Equation \ref{eqn:sparhowithmultipliers}, we get:
\begin{align}
\nonumber
\sparhov &= \onev + \Qv \frac{\piv^T \Qv - \muv^T \Qv}{\muv^T (\Qv \circ \Qv) - (\muv^T \Qv)^2} - \muv^T \Qv \frac{\piv^T \Qv - \muv^T \Qv}{\muv^T (\Qv \circ \Qv) - (\muv^T \Qv)^2} \\
\nonumber
\sparhov &= \onev + \frac{\Qv - \muv^T \Qv}{\muv^T (\Qv \circ \Qv) - (\muv^T \Qv)^2} (\piv^T \Qv - \muv^T \Qv)
\end{align}
Replacing the inner products with the expectations they correspond with gives us:
\begin{equation}
\sparhov = \onev + \frac{\Qv - \Ex_\mu[Q]}{\Ex_\mu[Q^2] - \Ex_\mu[Q]^2} (\Ex_\pi[Q] - \Ex_\mu[Q])
\end{equation}

\newpage
\section{$\lambda$-return Backward-view Derivation}

We start with the recursive definition of the  $\lambda$-return for Sarsa($\lambda$) \citep{precup2000pdis}, one of the simplest TD($\lambda$) algorithms for off-policy action-value estimation:
\begin{equation*}
    G_t^{\lambda} = R_{t+1} + \gamma \Big( \lambda \rho_{t+1} G_{t+1}^\lambda + (1 - \lambda) \rho_{t + 1} Q(S_{t+1}, A_{t+1}) \Big)
\end{equation*}
Adding the control variate terms to the discounted future rewards, we get:
\begin{align*}
    G_t^{\lambda} = R_{t+1} + \gamma \Big( &\lambda \rho_{t+1} G_{t+1}^\lambda + (1 - \lambda) \rho_{t + 1} Q(S_{t+1}, A_{t+1}) + \Ex_\pi [Q(S_{t+1}, \cdot)] - \rho_{t+1} Q(S_{t+1}, A_{t+1}) \Big) \\
    G_t^{\lambda} = R_{t+1} + \gamma \Big( &\lambda \big( \rho_{t+1} G_{t+1}^{\lambda} + \Ex_\pi [Q(S_{t+1}, \cdot)] - \rho_{t+1} Q(S_{t+1}, A_{t+1}) \big) + \\
    &(1 - \lambda) \big( \rho_{t+1} Q(S_{t+1}, A_{t+1}) + \Ex_\pi [Q(S_{t+1}, \cdot)] - \rho_{t+1} Q(S_{t+1}, A_{t+1}) \big) \Big) \\
    G_t^{\lambda} = R_{t+1} + \gamma \Big( &\lambda \big( \rho_{t+1} G_{t+1}^{\lambda} + \Ex_\pi [Q(S_{t+1}, \cdot)] - \rho_{t+1} Q(S_{t+1}, A_{t+1})\big) + (1 - \lambda) \Ex_\pi [Q(S_{t+1}, \cdot)] \Big)
\end{align*}
We can re-write this as a recursion of $G_t^\lambda - Q(S_t, A_t)$:
\begin{align*}
    G_t^{\lambda} &= R_{t+1} + \gamma \Big( \lambda \big( \rho_{t+1} G_{t+1}^{\lambda} + \Ex_\pi [Q(S_{t+1}, \cdot)] - \rho_{t+1} Q(S_{t+1}, A_{t+1})\big) + (1 - \lambda) \Ex_\pi [Q(S_{t+1}, \cdot)] \Big) \\
    G_t^{\lambda} &= R_{t+1} + \gamma \lambda \rho_{t+1} G_{t+1}^{\lambda} + \gamma \lambda \Ex_\pi [Q(S_{t+1}, \cdot)] - \gamma \lambda \rho_{t+1} Q(S_{t+1}, A_{t+1}) + \gamma \Ex_\pi [Q(S_{t+1}, \cdot)]  - \gamma \lambda \Ex_\pi [Q(S_{t+1}, \cdot)] \\
    G_t^{\lambda} &= R_{t+1} + \gamma \Ex_\pi [Q(S_{t+1}, \cdot)] + \gamma \lambda \rho_{t+1} G_{t+1}^{\lambda}- \gamma \lambda \rho_{t+1} Q(S_{t+1}, A_{t+1}) \\
    G_t^{\lambda} &= R_{t+1} + \gamma \Ex_\pi [Q(S_{t+1}, \cdot)] + \gamma \lambda \rho_{t+1} \big( G_{t+1}^{\lambda} -  Q(S_{t+1}, A_{t+1}) \big) \\
    G_t^{\lambda} - Q(S_t, A_t) &= R_{t+1} + \gamma \Ex_\pi [Q(S_{t+1}, \cdot)] - Q(S_t, A_t) + \gamma \lambda \rho_{t+1} \big( G_{t+1}^{\lambda} -  Q(S_{t+1}, A_{t+1}) \big) \\
\end{align*}
From here, we get a sum of one-step Expected Sarsa TD errors, discounted by $\gamma\lambda\rho_t$:
\begin{equation*}
    G_t^{\lambda} = Q(S_t, A_t) + \sum_{k=t}^{\infty}{\big(R_{k+1} + \gamma \Ex_\pi [Q(S_{k+1}, \cdot)] - Q(S_k, A_k) \big) \prod_{i=t+1}^{k}{\gamma \lambda \rho_i}}
\end{equation*}
To provide intuition on the effect of the control variate, following similar steps with the Sarsa($\lambda$)'s $\lambda$-return gives:
\begin{align*}
    G_t^{\lambda} &= R_{t+1} + \gamma \Big( \lambda \rho_{t+1} G_{t+1}^\lambda + (1 - \lambda) \rho_{t + 1} Q(S_{t+1}, A_{t+1}) \Big) \\
    G_t^{\lambda} &= R_{t+1} + \gamma \lambda \rho_{t+1} G_{t+1}^\lambda + \gamma \rho_{t + 1} Q(S_{t+1}, A_{t+1}) - \gamma \lambda \rho_{t + 1} Q(S_{t+1}, A_{t+1}) \\
    G_t^{\lambda} &= R_{t+1} + \gamma \rho_{t + 1} Q(S_{t+1}, A_{t+1})  + \gamma \lambda \rho_{t+1} G_{t+1}^\lambda - \gamma \lambda \rho_{t+1} Q(S_{t+1}, A_{t+1}) \\
    G_t^{\lambda} &= R_{t+1} + \gamma \rho_{t + 1} Q(S_{t+1}, A_{t+1})  + \gamma \lambda \rho_{t+1} \big( G_{t+1}^\lambda - Q(S_{t+1}, A_{t+1}) \big) \\
    G_t^{\lambda} - Q(S_t, A_t) &= R_{t+1} + \gamma \rho_{t + 1} Q(S_{t+1}, A_{t+1}) - Q(S_t, A_t) + \gamma \lambda \rho_{t+1} \big( G_{t+1}^\lambda - Q(S_{t+1}, A_{t+1}) \big) \\
    G_t^{\lambda} &= Q(S_t, A_t) + \sum_{k=t}^{\infty}{\big(R_{k+1} + \gamma \rho_{k+1} Q(S_{k+1}, A_{k+1}) - Q(S_k, A_k) \big) \prod_{i=t+1}^{k}{\gamma \lambda \rho_i}}
\end{align*}
We see that the control variate is responsible for the use of one-step Expected Sarsa TD errors being weighted by the eligibility trace, as opposed to those of one-step Sarsa with importance sampling. It's clear that computing the complete expectation is typically favorable over relying on averaging out importance weighted-samples in the bootstrap target, and can be expected to result in increased variance on the order of what is presented in Figure \ref{fig:bandit_results}.

\newpage
\section{Additional Experimental Details}
\label{app:expdetails}



\subsection{Path World Details}

Given actions $a \in \{0, 1, 2, \dots, |\mathcal{A}|\}$, rewards were specified by $r(s,a) = \frac{1 + a}{ |\mathcal{A}|}$.

The scope of the parameter sweep is detailed below:

\begin{center}
    \begin{tabular}{ |c|c| } 
     \hline
     \textbf{Parameter} & \textbf{Value(s)} \\
     \hline
     Discount Factor $\gamma$ & $1.0$ \\
     \hline
     Action Space Size $|\mathcal{A}|$ & $8, 32$ \\
     \hline
     Step Size $\alpha$ & $0.1, 0.2, 0.3, 0.4, 0.5, 0.6, 0.7, 0.8, 0.9, 1.0$ \\
     \hline
     Trace Decay $\lambda$ & $0.5, 0.75, 0.875$ \\
     \hline
    \end{tabular}
\end{center}

\subsection{5 $\times$ 5 Grid World Details}

The scope of the parameter sweep for the tabular results are detailed below:

\begin{center}
    \begin{tabular}{ |c|c| } 
     \hline
     \textbf{Parameter} & \textbf{Value(s)} \\
     \hline
     Discount Factor $\gamma$ & $1.0$ \\
     \hline
     Action Space Size $|\mathcal{A}|$ & $4$ \\
     \hline
     Action Commitment $\epsilon_\pi$ & $0.5$ \\
     \hline
     Action Commitment $\epsilon_\mu$ & $1.0$ \\
     \hline
     Step Size $\alpha$ & $0.1, 0.2, 0.3, 0.4, 0.5, 0.6, 0.7, 0.8, 0.9, 1.0$ \\
     \hline
     Trace Decay $\lambda$ & $0.5, 0.75, 0.875, 0.9375, 0.96875, 0.984375, 0.9921875$ \\
     \hline
    \end{tabular}
\end{center}

The scope of the parameter sweep for the linear function approximation results are detailed below:

\begin{center}
    \begin{tabular}{ |c|c| } 
     \hline
     \textbf{Parameter} & \textbf{Value(s)} \\
     \hline
     Discount Factor $\gamma$ & $1.0$ \\
     \hline
     Action Space Size $|\mathcal{A}|$ & $8$ \\
     \hline
     Action Commitment $\epsilon_\pi$ & $0.5$ \\
     \hline
     Action Commitment $\epsilon_\mu$ & $0.5$ \\
     \hline
     Step Size $\alpha$ & $0.1, 0.2, 0.3, 0.4, 0.5, 0.6, 0.7, 0.8, 0.9, 1.0$ \\
     \hline
     Trace Decay $\lambda$ & $0.5, 0.75, 0.875, 0.9375$ \\
     \hline
    \end{tabular}
\end{center}

\subsection{Acrobot-v1 Details}

The scope of the parameter sweep is detailed below:

\begin{center}
    \begin{tabular}{ |c|c| } 
     \hline
     \textbf{Parameter} & \textbf{Value(s)} \\
     \hline
     Discount Factor $\gamma$ & $0.99$ \\
     \hline
     Time Limit & $500$ \\
     \hline
     Hidden Layers & $2$ \\
     \hline
     Hidden Layer Width & $32, 64, 128, 256, 512$ \\
     \hline
     Step Size $\alpha$ & $10^{-4.5}, 10^{-4}, 10^{-3.5}, 10^{-3}$ \\
     \hline
     Trace Decay $\lambda$ & $0.5, 0.75, 0.875, 0.9375$ \\
     \hline
     Initial $\epsilon$ & $1.0$ \\
     \hline
     Final $\epsilon$ & $0.1$ \\
     \hline
     Linear $\epsilon$ Decay Steps & $25000$ \\
     \hline
    \end{tabular}
\end{center}

\newpage
\section{Additional Experimental Results}
\label{app:moreexps}

\subsection{Online Bandit Results}

Below we consider the random bandit instances in the case where values are being learned, as the results presented in Figure \ref{fig:bandit_results} compute the relevant metrics in closed form given the true action-values. In addition to importance sampling, Sparho, and their clipped versions, we further evaluate regression importance sampling \citep{hanna2018ris} applied to both the importance sampling and Sparho estimators. We fix $|\mathcal{A}| = 8, \alpha = 0.01$ and note that the trends are reasonably consistent across the parameter space due to the simplicity of the problem. Averaged over 100 runs, we measure the absolute value error across 1,000 steps.

\begin{figure}[ht]
    \centering
    \includegraphics[width=0.5\textwidth]{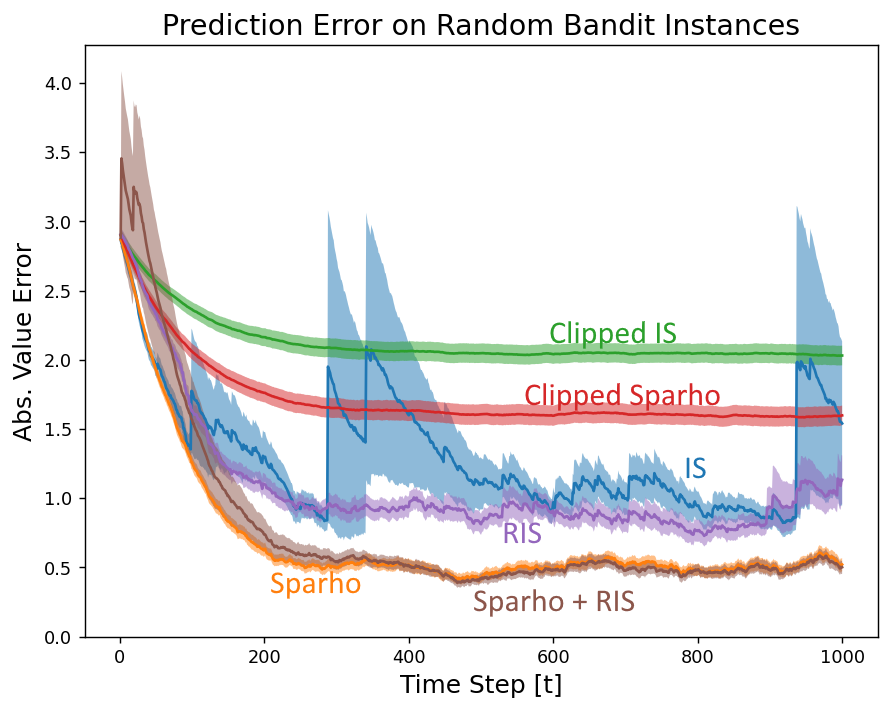}
    \caption{Absolute value error for various estimators across randomly generated bandit instances. Results are averaged over 100 instances, and shaded regions represent one standard error.}
    \label{fig:bandit_online_results}
\end{figure}

We see that even when action-values are being estimated, the Sparho estimator has substantially lower variance than the importance sampling counterpart for all variants tested (original, clipped, RIS). Given the use of a fixed step-size, this further leads to lower final value error as there is less variability to average out. We see general decrease in variance with the use of an estimated behavior policy for both RIS estimators. We further observe substantial bias in the solution of the clipped estimators, though the clipped Sparho estimator exhibited less of it.

\subsection{Emphatic TD($\lambda$) in the 5 $\times$ 5 Grid World}

\begin{figure}[ht]
    \centering
    \includegraphics[width=0.45\textwidth]{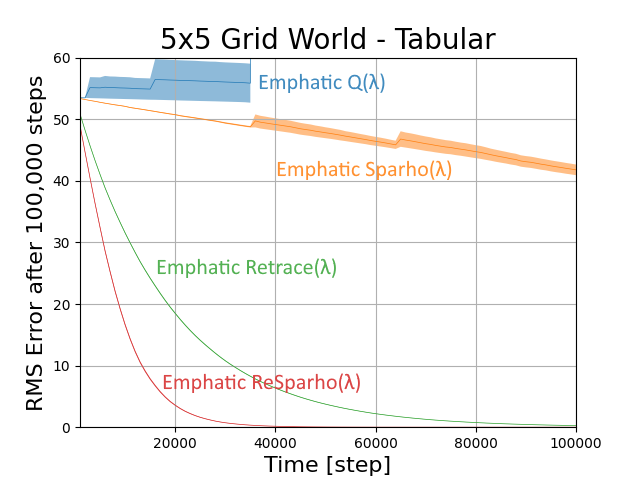}
    \caption{Results in the 5 $\times$ 5 Grid World environment with a tabular representation for Emphatic Q($\lambda$) with various choices of importance weights. Results are averaged over 100 runs, and shaded regions represent one standard error.}
    \label{fig:gw4_etd}
\end{figure}

Here we evaluate the substitution of the Sparho estimator in Emphatic TD($\lambda$). Specifically, we evaluate EQ($\lambda$) with importance sampling, Sparho, as well as their clipped versions, in the 5 $\times$ 5 Grid World environment with a tabular representation. We assume uniform interest $i(s) = 1, \forall s \in \mathcal{S}$, perform a sweep over step size $\alpha$ and trace-decay rate $\lambda$ (detailed below), and run each instance for 100,000 steps. Results are averaged over 100 runs, and the best parameter settings among the sweep in terms of final RMS error are shown in Figure \ref{fig:gw4_etd}.

\begin{center}
    \begin{tabular}{ |c|c| } 
     \hline
     \textbf{Parameter} & \textbf{Value(s)} \\
     \hline
     Discount Factor $\gamma$ & $1.0$ \\
     \hline
     Action Space Size $|\mathcal{A}|$ & $4$ \\
     \hline
     Action Commitment $\epsilon_\pi$ & $0.5$ \\
     \hline
     Action Commitment $\epsilon_\mu$ & $1.0$ \\
     \hline
     Step Size $\alpha$ & $0.0001, 0.0005, 0.001, 0.005, 0.01, 0.05, 0.1$ \\
     \hline
     Trace Decay $\lambda$ & $0.5, 0.75, 0.875, 0.9375, 0.96875$ \\
     \hline
    \end{tabular}
\end{center}

We observe that despite the relative simplicity of the problem (a small, structured grid world), Emphatic Q($\lambda$) with importance sampling doesn't reliably learn. Substituting $\rho_t \rightarrow \sparho_t$ improves performance, but slowly progresses due to the necessity of a small step size to handle the base algorithm's increased variance. Clipping both estimators enables substantially better learning progress, but notably much slower than their non-emphatic counterparts in Figure \ref{fig:gw4_results}.

\newpage
\section{Alternate Objectives}
\label{app:altobjs}

Below we provide importance weight derivations for alternate optimization objectives.

\subsection{Minimum $\ell$2 distance to 1}

The following is the Sparho objective but without the constraint on the expected importance weight being 1. This can have lower variance importance weights which are often close to 1, but may arbitrarily attenuate or amplify a sequence in the multi-step case.

\begin{equation*}
\mathcal{L}(\sparho, \lambda) = \sum_{a}{\mu_a(\sparho_a - 1)^2} + \lambda \Big( \sum_{a}{\pi_a Q_a} - \sum_{a}{\mu_a \sparho_a Q_a} \Big)
\end{equation*}

\begin{equation*}
\mathcal{L}(\sparhov, \lambda) = (\sparhov - 1)^T D_{\mu} (\sparhov - 1) + \lambda_0 \big( \piv^T \Qv - (D_\mu \Qv)^T \sparhov \big)
\end{equation*}

\begin{align*}
\nabla \mathcal{L}(\sparhov, \lambda) = 0 &= 2 D_\mu \sparhov - 2 \muv - D_\mu \Qv \lambda \\
0 &= D_\mu \sparhov - \muv - D_\mu \Qv \frac{\lambda}{2} \\
0 &= D^{-1}_\mu D_\mu \sparhov - D^{-1}_\mu \muv - D^{-1}_\mu D_\mu \Qv \frac{\lambda}{2} \\
0 &= \sparhov - \onev - \Qv \frac{\lambda}{2} \\
\sparhov &= \onev + \Qv \frac{\lambda}{2} \\
(D_\mu \Qv)^T \sparhov &= (D_\mu \Qv)^T \onev + (D_\mu \Qv)^T \Qv \frac{\lambda}{2} \\
\piv^T \Qv &= \muv^T \Qv + \muv^T (\Qv \circ \Qv) \frac{\lambda}{2} \\
\frac{\lambda}{2} &= \frac{\piv^T \Qv - \muv^T \Qv}{\muv^T (\Qv \circ \Qv)} \\
\sparhov &= \onev + \frac{\Qv}{\muv^T (\Qv \circ \Qv)}(\piv^T \Qv - \muv^T \Qv)
\end{align*}
\begin{equation*}
\sparhov = \onev + \frac{\Qv}{\Ex_\mu[Q^2]} (\Ex_\pi[Q] - \Ex_\mu[Q])
\end{equation*}

\subsection{Minimum $\ell$2 distance to $c$}

This is a generalization of C.1 to have the weights tend toward a specific value $c$, but again without a hard constraint on the expected importance weight.

\begin{equation*}
\mathcal{L}(\sparho, \lambda) = \sum_{a}{\mu_a(\sparho_a - c)^2} + \lambda \Big( \sum_{a}{\pi_a Q_a} - \sum_{a}{\mu_a \sparho_a Q_a} \Big)
\end{equation*}

\begin{equation*}
\mathcal{L}(\sparhov, \lambda) = (\sparhov - c)^T D_{\mu} (\sparhov - c) + \lambda_0 \big( \piv^T \Qv - (D_\mu \Qv)^T \sparhov \big)
\end{equation*}

\begin{align*}
\nabla \mathcal{L}(\sparhov, \lambda) = 0 &= 2 D_\mu \sparhov - 2 c \muv - D_\mu \Qv \lambda \\
0 &= D_\mu \sparhov - c \muv - D_\mu \Qv \frac{\lambda}{2} \\
0 &= D^{-1}_\mu D_\mu \sparhov - D^{-1}_\mu c \muv - D^{-1}_\mu D_\mu \Qv \frac{\lambda}{2} \\
0 &= \sparhov - \cv - \Qv \frac{\lambda}{2} \\
\sparhov &= \cv + \Qv \frac{\lambda}{2} \\
(D_\mu \Qv)^T \sparhov &= (D_\mu \Qv)^T \cv + (D_\mu \Qv)^T \Qv \frac{\lambda}{2} \\
\piv^T \Qv &= c \muv^T \Qv + \muv^T (\Qv \circ \Qv) \frac{\lambda}{2} \\
\frac{\lambda}{2} &= \frac{\piv^T \Qv - c \muv^T \Qv}{\muv^T (\Qv \circ \Qv)} \\
\sparhov &= \cv + \frac{\Qv}{\muv^T (\Qv \circ \Qv)}(\piv^T \Qv - c \muv^T \Qv)
\end{align*}
\begin{equation*}
\sparhov = \cv + \frac{\Qv}{\Ex_\mu[Q^2]} (\Ex_\pi[Q] - c \Ex_\mu[Q])
\end{equation*}

\subsection{Minimum variance around constrained length $c$}

This is a generalization of the minimum-variance Sparho objective, but with a variable expected importance weight magnitude. One possible application of this is to absorb the discount factor into the importance weights to simultaneously discount and correct the distribution of a sample.

\begin{equation*}
\mathcal{L}(\sparho, \lambda_0, \lambda_1) = \sum_{a}{\mu_a(\sparho_a - c)^2} + \lambda_0 \Big( \sum_{a}{\pi_a Q_a} - \sum_{a}{\mu_a \sparho_a Q_a} \Big) + \lambda_1 \Big( c - \sum_{a}{\mu_a\sparho_a} \Big)
\end{equation*}
\begin{equation*}
\mathcal{L}(\sparhov, \lambda_0, \lambda_1) = (\sparhov - c)^T D_{\mu} (\sparhov - c) + \lambda_0 \big( \piv^T \Qv - (D_\mu \Qv)^T \sparhov \big) + \lambda_1 (c - (D_\mu \onev)^T \sparhov)
\end{equation*}
\begin{align*}
\nabla \mathcal{L}(\sparhov, \lambda_0, \lambda_1) = 0 &= 2 D_\mu \sparhov - 2 c \muv - D_\mu \Qv \lambda_0 - D_\mu \onev \lambda_1 \\
0 &= D_\mu \sparhov - c \muv - D_\mu \Qv \frac{\lambda_0}{2} - D_\mu \onev \frac{\lambda_1}{2} \\
0 &= D_\mu^{-1} D_\mu \sparhov - D_\mu^{-1} c \muv - D_\mu^{-1} D_\mu \Qv \frac{\lambda_0}{2} - D_\mu^{-1} D_\mu \onev \frac{\lambda_1}{2} \\
0 &= \sparhov - \cv - \Qv \frac{\lambda_0}{2} - \onev \frac{\lambda_1}{2} \\
\sparhov &= \cv + \Qv \frac{\lambda_0}{2} + \onev \frac{\lambda_1}{2}
\end{align*}
\begin{align*}
(D_\mu \Qv)^T \sparhov &= (D_\mu \Qv)^T \cv + (D_\mu \Qv)^T \Qv \frac{\lambda_0}{2} + (D_\mu \Qv)^T \onev \frac{\lambda_1}{2} \\
(D_\mu \Qv)^T \sparhov - (D_\mu \Qv)^T \cv &= (D_\mu \Qv)^T \Qv \frac{\lambda_0}{2} + (D_\mu \Qv)^T \onev \frac{\lambda_1}{2} \\
\piv^T \Qv - c \muv^T \Qv &= \muv^T (\Qv \circ \Qv) \frac{\lambda_0}{2} + \muv^T \Qv \frac{\lambda_1}{2}
\end{align*}
\begin{align*}
(D_\mu \onev)^T \sparhov &= (D_\mu \onev)^T \cv + (D_\mu \onev)^T \Qv \frac{\lambda_0}{2} + (D_\mu \onev)^T \onev \frac{\lambda_1}{2} \\
(D_\mu \onev)^T \sparhov - (D_\mu \onev)^T \cv &= (D_\mu \onev)^T \Qv \frac{\lambda_0}{2} + (D_\mu \onev)^T \onev \frac{\lambda_1}{2} \\
0 &= \muv^T \Qv \frac{\lambda_0}{2} + \frac{\lambda_1}{2}
\end{align*}
\begin{align*}
\begin{bmatrix}
\muv^T (\Qv \circ \Qv) & \muv^T \Qv \\
\muv^T \Qv & 1
\end{bmatrix}
\begin{bmatrix}
\frac{\lambda_0}{2} \\
\frac{\lambda_1}{2}
\end{bmatrix} &= 
\begin{bmatrix}
\piv^T \Qv - c \muv^T \Qv \\
0 \\
\end{bmatrix} \\
\begin{bmatrix}
\frac{\lambda_0}{2} \\
\frac{\lambda_1}{2}
\end{bmatrix} &=
\begin{bmatrix}
\muv^T (\Qv \circ \Qv) & \muv^T \Qv \\
\muv^T \Qv & 1
\end{bmatrix}^{-1}
\begin{bmatrix}
\piv^T \Qv - c \muv^T \Qv \\
0 \\
\end{bmatrix} \\
\frac{\lambda_0}{2} &= \frac{\piv^T \Qv - c \muv^T \Qv}{\muv^T (\Qv \circ \Qv) - (\muv^T \Qv)^2} \\
\frac{\lambda_1}{2} &= -\muv^T \Qv \frac{\piv^T \Qv - c \muv^T \Qv}{\muv^T (\Qv \circ \Qv) - (\muv^T \Qv)^2}
\end{align*}
\begin{align*}
\sparhov &= \cv + \Qv \frac{\piv^T \Qv - c \muv^T \Qv}{\muv^T (\Qv \circ \Qv) - (\muv^T \Qv)^2} - \muv^T \Qv \frac{\piv^T \Qv - c \muv^T \Qv}{\muv^T (\Qv \circ \Qv) - (\muv^T \Qv)^2} \\
\sparhov &= \cv + \frac{\Qv - \muv^T \Qv}{\muv^T (\Qv \circ \Qv) - (\muv^T \Qv)^2} (\piv^T \Qv - c \muv^T \Qv)
\end{align*}
\begin{equation*}
\sparhov = \cv + \frac{\Qv - \Ex_\mu[Q]}{\Ex_\mu[Q^2] - \Ex_\mu[Q]^2} (\Ex_\pi[Q] - c \Ex_\mu[Q])
\end{equation*}

\subsection{Minimum variance of $\sparho Q$}

This objective aims to minimize the variance of $\sparho Q$. Such an objective has intuitive appeal in minimizing the variance of the full update target, as opposed to just the importance weight. However, despite the focus on value-awareness, such an objective may be too reliant on the specific value estimates, and very sensitive to deviations in them. This may lead to substantially more bias, and even an increase in variance should the value estimates not accurately reflect the return.

Let $V = \mathbb{E}_\pi[Q] = \sum_a{\pi_a Q_a}$:

\begin{equation*}
\mathcal{L}(\sparho, \lambda_0, \lambda_1) = \sum_{a}{\mu_a(\sparho_a Q_a - V)^2} + \lambda_0 \Big( V - \sum_{a}{\mu_a \sparho_a Q_a} \Big) + \lambda_1 \Big( 1 - \sum_{a}{\mu_a\sparho_a} \Big)
\end{equation*}
\begin{equation*}
\mathcal{L}(\sparhov, \lambda_0, \lambda_1) = (D_Q \sparhov - V)^T D_{\mu} (D_Q \sparhov - V) + \lambda_0 \big( V - (D_\mu \Qv)^T \sparhov \big) + \lambda_1 (1 - (D_\mu \onev)^T \sparhov)
\end{equation*}
\begin{align*}
\nabla \mathcal{L}(\sparhov, \lambda_0, \lambda_1) = 0 &= 2 D_\mu D_Q D_Q \sparhov - 2 D_\mu \Qv V - D_\mu \Qv \lambda_0 - D_\mu \onev \lambda_1 \\
0 &= D_\mu D_Q D_Q \sparhov - D_\mu \Qv V - D_\mu \Qv \frac{\lambda_0}{2} - D_\mu \onev \frac{\lambda_1}{2} \\
0 &= D_Q^{-1} D_Q^{-1} D_\mu^{-1} D_\mu D_Q D_Q \sparhov - D_Q^{-1} D_Q^{-1} D_\mu^{-1} D_\mu \Qv V - D_Q^{-1} D_Q^{-1} D_\mu^{-1} D_\mu \Qv \frac{\lambda_0}{2} - D_Q^{-1} D_Q^{-1} D_\mu^{-1} D_\mu \onev \frac{\lambda_1}{2} \\
0 &= \sparhov - D_Q^{-1} \onev V - D_Q^{-1} \onev \frac{\lambda_0}{2} - D_Q^{-1} D_Q^{-1} \onev \frac{\lambda_1}{2} \\
\sparhov &= D_Q^{-1} \onev V + D_Q^{-1} \onev \frac{\lambda_0}{2} + D_Q^{-1} D_Q^{-1} \onev \frac{\lambda_1}{2}
\end{align*}
\begin{align*}
(D_\mu \Qv)^T \sparhov &= (D_\mu \Qv)^T D_Q^{-1} \onev V + (D_\mu \Qv)^T D_Q^{-1} \onev \frac{\lambda_0}{2} + (D_\mu \Qv)^T D_Q^{-1} D_Q^{-1} \onev \frac{\lambda_1}{2} \\(D_\mu \Qv)^T \sparhov - (D_\mu \Qv)^T D_Q^{-1} \onev V &= (D_\mu \Qv)^T D_Q^{-1} \onev \frac{\lambda_0}{2} + (D_\mu \Qv)^T D_Q^{-1} D_Q^{-1} \onev \frac{\lambda_1}{2} \\
0 &= \frac{\lambda_0}{2} + \muv^T D_\mu^{-1} \onev \frac{\lambda_1}{2}
\end{align*}
\begin{align*}
(D_\mu \onev)^T \sparhov &= (D_\mu \onev)^T D_Q^{-1} \onev V + (D_\mu \onev)^T D_Q^{-1} \onev \frac{\lambda_0}{2} + (D_\mu \onev)^T D_Q^{-1} D_Q^{-1} \onev \frac{\lambda_1}{2} \\
(D_\mu \onev)^T \sparhov - (D_\mu \onev)^T D_Q^{-1} \onev V &= (D_\mu \onev)^T D_Q^{-1} \onev \frac{\lambda_0}{2} + (D_\mu \onev)^T D_Q^{-1} D_Q^{-1} \onev \frac{\lambda_1}{2} \\
1 - \muv^T D_Q^{-1} \onev V &= \muv^T D_Q^{-1} \onev \frac{\lambda_0}{2} + \muv^T D_Q^{-1} D_Q^{-1} \onev \frac{\lambda_1}{2}
\end{align*}
\begin{align*}
\begin{bmatrix}
1 & \muv^T D_Q^{-1} \onev \\
\muv^T D_Q^{-1} \onev & \muv^T D_Q^{-1} D_Q^{-1} \onev
\end{bmatrix}
\begin{bmatrix}
\frac{\lambda_0}{2} \\
\frac{\lambda_1}{2}
\end{bmatrix} &= 
\begin{bmatrix}
0 \\
1 - \muv^T D_Q^{-1} \onev V \\
\end{bmatrix} \\
\begin{bmatrix}
\frac{\lambda_0}{2} \\
\frac{\lambda_1}{2}
\end{bmatrix} &=
\begin{bmatrix}
1 & \muv^T D_Q^{-1} \onev \\
\muv^T D_Q^{-1} \onev & \muv^T D_Q^{-1} D_Q^{-1} \onev
\end{bmatrix}^{-1}
\begin{bmatrix}
0 \\
1 - \muv^T D_Q^{-1} \onev V \\
\end{bmatrix} \\
\frac{\lambda_0}{2} &= \frac{(\muv^T D_Q^{-1} \onev)^2 V - \muv^T D_Q^{-1} \onev}{\muv^T D_Q^{-1} D_Q^{-1} \onev - (\muv^T D_Q^{-1} \onev)^2} \\
\frac{\lambda_1}{2} &= \frac{1 - \muv^T D_Q^{-1} \onev V}{\muv^T D_Q^{-1} D_Q^{-1} \onev - (\muv^T D_Q^{-1} \onev)^2}
\end{align*}
\begin{align*}
\sparhov &= D_Q^{-1} \onev V + D_Q^{-1} \onev \frac{(\muv^T D_Q^{-1} \onev)^2 V - \muv^T D_Q^{-1} \onev}{\muv^T D_Q^{-1} D_Q^{-1} \onev - (\muv^T D_Q^{-1} \onev)^2} + D_Q^{-1} D_Q^{-1} \onev \frac{1 - \muv^T D_Q^{-1} \onev V}{\muv^T D_Q^{-1} D_Q^{-1} \onev - (\muv^T D_Q^{-1} \onev)^2}
\end{align*}
\begin{equation*}
\sparho_a = \frac{V}{Q_a} + \frac{\frac{1}{Q_a}(\Ex_\mu[\frac{1}{Q}]^2 V - \Ex_\mu[\frac{1}{Q}]) + \frac{1}{Q_a^2}(1 - \Ex_\mu[\frac{1}{Q}] V)}{\Ex_\mu[\frac{1}{Q^2}] - \Ex_\mu[\frac{1}{Q}]^2}
\end{equation*}

\end{document}

%% file: math_commands.tex
\usepackage{amsmath,amsfonts,bm}

\def\eqref#1{equation~\ref{#1}}

\def\1{\bm{1}}

\DeclareMathAlphabet{\mathsfit}{\encodingdefault}{\sfdefault}{m}{sl}
\SetMathAlphabet{\mathsfit}{bold}{\encodingdefault}{\sfdefault}{bx}{n}

